\documentclass[conference]{IEEEtran}
\IEEEoverridecommandlockouts
\usepackage{cite}
\usepackage{amsmath,amssymb,amsfonts}
\usepackage{float}
\usepackage{graphicx}
\usepackage{textcomp}
\usepackage{xcolor}
\usepackage{subfig}
\usepackage{booktabs}
\usepackage{multirow}

\usepackage{algorithm}
\usepackage{algpseudocode}

\def\BibTeX{{\rm B\kern-.05em{\sc i\kern-.025em b}\kern-.08em
    T\kern-.1667em\lower.7ex\hbox{E}\kern-.125emX}}
\begin{document}

\title{Comparative Analysis of UAV Path Planning Algorithms for Efficient Navigation in Urban 3D Environments\\

}

\author{\IEEEauthorblockN{1\textsuperscript{st} Hichem Cheriet}
\IEEEauthorblockA{\textit{dept. of Computer Science} \\
\textit{Université d'USTO Mohamed Boudiaf}\\
Oran, Algeria \\
hichem.cheriet@univ-usto.dz}
\and
\IEEEauthorblockN{2\textsuperscript{nd} Khellat Kihel Badra}
\IEEEauthorblockA{\textit{dept. of Economics} \\
\textit{Oran2 Mohamed BenAhmed University}\\
Oran, Algeria \\
khellat\_badra@yahoo.fr}
\and
\IEEEauthorblockN{3\textsuperscript{rd} Chouraqui Samira}
\IEEEauthorblockA{\textit{dept. of Computer Science} \\
\textit{Université d'USTO Mohamed Boudiaf}\\
Oran, Algeria \\
samirachouraqui178@gmail.com}

}

\maketitle

\begin{abstract}
The most crucial challenges for UAVs are planning paths and avoiding obstacles in their way. In recent years, a wide variety of path-planning algorithms have been developed. These algorithms have successfully solved path-planning problems; however, they suffer from multiple challenges and limitations. To test the effectiveness and efficiency of three widely used algorithms, namely A*, RRT*, and Particle Swarm Optimization (PSO), this paper conducts extensive experiments in 3D urban city environments cluttered with obstacles. Three experiments were designed with two scenarios each to test the aforementioned algorithms. These experiments consider different city map sizes, different altitudes, and varying obstacle densities and sizes in the environment. According to the experimental results, the A* algorithm outperforms the others in both computation efficiency and path quality. PSO is especially suitable for tight turns and dense environments, and RRT* offers a balance and works well across all experiments due to its randomized approach to finding solutions.

\end{abstract}


\section{Introduction}

Unmanned aerial vehicles (UAVs) with different shapes and types are widely utilized in many fields, such as military \cite{b2}, photography, search and rescue  \cite{b1}, and agriculture \cite{b8}. The path-finding problem has been a key area of research these years for effectively navigating terrains, successfully completing missions, and arriving at target positions. Path planning, in general, can be separated into two principal approaches. The first approach is used when the environment is already known, and the threats maintain their position without movement, such as buildings, bridges, and traffic lights. This is called static path planning, referring to the static environment. Path planning for this type is calculated before the mission starts. The second type is called dynamic path planning, referring to the dynamic environment where obstacles change their positions, such as vehicles, birds, humans, etc. Dynamic path planning requires an online path calculator in the UAV to adapt to any environmental changes effectively. The choice between these two approaches depends on the mission requirement.

Path planning algorithms can be categorized into four main approaches. Firstly, graph-based search algorithms treat the environment as a graph. In this approach, nodes represent specific locations or states, and edges represent possible transitions or movements between these locations, such as A* \cite{b4}, Dijkstra's algorithm \cite{b3}, D* \cite{b31}, and JPS (Jump Point Search) \cite{b32}. The second approach comprises sampling-based techniques, including Probabilistic Roadmaps  (PRM) \cite{b5} and rapidly exploring random trees (RRTs) \cite{b6}. These algorithms randomize the search space into samples and construct paths based on these samples. The third approach involves Artificial Potential Fields \cite{b33}, where attractive forces guide the UAV (Unmanned Aerial Vehicle) toward the goal. Recently, bio-inspired algorithms have been utilized to address path-planning challenges. These algorithms draw inspiration from biological phenomena like animal behavior and rely on a predefined cost function. Examples include ant colony optimization (ACO) \cite{b34}, particle swarm optimization (PSO) \cite{b9}, and grey wolf optimization (GWO) \cite{b10}. Each of these methods has its advantages. However, there are still many challenges associated with using these approaches, and no single algorithm is suitable for all scenarios. For example, graph-based methods are computationally expensive, especially in 3D spaces and large environments \cite{b35}. Bio-inspired methods require parameter tuning and may necessitate running simulations for numerous generations to find a good solution \cite{b36}. APF (Artificial Potential Field) methods can be easily trapped in local minima, particularly when navigating close to obstacles \cite{b37}. Sampling-based methods typically provide a series of waypoints rather than a smooth, continuous trajectory, which can be challenging for UAVs with complex dynamics that require precise motion control \cite{b38}.
A number of hybrid approaches have been developed for UAV path planning to address the limitations of conventional methods, including PSO-APF \cite{psoapf}, APPATT \cite{appatt}, and the Tangent A* planner \cite{tangentastar}, among others.

Among these algorithms, RRT* \cite{b39}, A*, and PSO are classical methods frequently employed in various path-planning scenarios. This paper specifically addresses the UAV path planning challenge within an urban 3D environment. It compares and analyzes the advantages of the three classical algorithms under different flying conditions. To effectively choose the best path planning algorithm, six different scenarios are designed and applied, considering environment size, various obstacle sizes, densities, and shapes, as well as different altitudes between the start and the target positions.

The structure of the paper is as follows: Section 2 reviews related comparison studies in the field. Section 3 presents a concise summary of the three most commonly used algorithms: A*, RRt*, and PSO. Section 5 details the simulation and Analysis. Section 6 presents the conclusion.

\section{ Related Works}

Several studies have compared these algorithms in the field of path planning. For instance, Zakir et al.\cite{b45} evaluate an improved Dijkstra algorithm combined with a Voronoi diagram and the A* algorithm to minimize threat collisions and fuel consumption. The results found that the Voronoi-Dijkstra approach is effective in stable environments, while A* finds paths between nodes using an effective heuristic. Kevin et al.\cite{b46} compare Dijkstra’s and waypoint generation algorithms. The study evaluates both algorithms across various scenarios, and the results show that Dijkstra offers precise solutions, but its performance varies depending on the cost function. However, The heuristic waypoint generation method provides a simple and faster search but sometimes has collision. Zammit et al. \cite{b47} evaluate A* and RRT algorithms for 3D UAV path planning in complex scenarios with vertical and horizontal obstacles. Their study shows that A* consistently offers shorter path lengths and faster generation times than RRT, making it the more efficient choice for online 3D navigation. Ziang et al.\cite{b48} compare three path planning algorithms: A*, RRT, and ACO  for UAV path planning in complex 3D city environments. The study finds that A* performs consistently well, ACO is best for large-scale scenarios with significant height differences, and RRT generally fails due to its randomness technique. Hu et al. \cite{b49} compare Genetic Algorithms (GA), Ant Colony Optimization (ACO), and Particle Swarm Optimization (PSO) for optimizing mission performance for UAVs. The experimental findings show that GA is better in complex environments, ACO is effective for multiple nodes, and PSO is best for real-time tasks.

\section{ The Three Utilized Algorithms Description}
In this section, the paper briefly explains the algorithms, namely A*, RRT*, and PSO.

\subsection{A* Algorithm}
A* is a graph-based algorithm that finds a path between two nodes in grids or graphs. It is a variant of the Dijkstra algorithm but uses a heuristic function to guide the search and ensure the minimization of the path length. A* usually solves path-finding problems and is often used in UAV missions \cite{b40}. To effectively explore the entire nodes, A* uses two sets: an open set and a closed set. The open set is used to store unexplored nodes, and the closed set is used for the explored ones. The heuristic function is used to calculate and define the chosen path and ensures the choice of the best one. The function is defined as follows:

\begin{equation}
\label{f_value}
    f(n) = g(n) + h(n)
\end{equation}

The equation uses $g(n)$ for the actual cost from the start to node $n$ and $h(n)$ for the estimated cost to the goal. The heuristic $h(n)$ must be admissible, and the true cost must never be overestimated. \\

To effectively plan a path in a specific environment, the algorithm first initializes two empty lists: an $open list$ and a $closed list$. The starting node is then added to the $open list$. For each node, the algorithm calculates its f value using Equation \ref{f_value}. Next, the algorithm repeatedly selects the node with the lowest $f$ value from the $open list$. It checks if this node is the goal; if so, the path is found and reconstructed from the goal to the start. Otherwise, the node is moved to the $closed list$, and its neighbors are evaluated. This process continues until the goal is reached or the $open list$ becomes empty.

\begin{figure}
    \centering
    \includegraphics[width=1\linewidth]{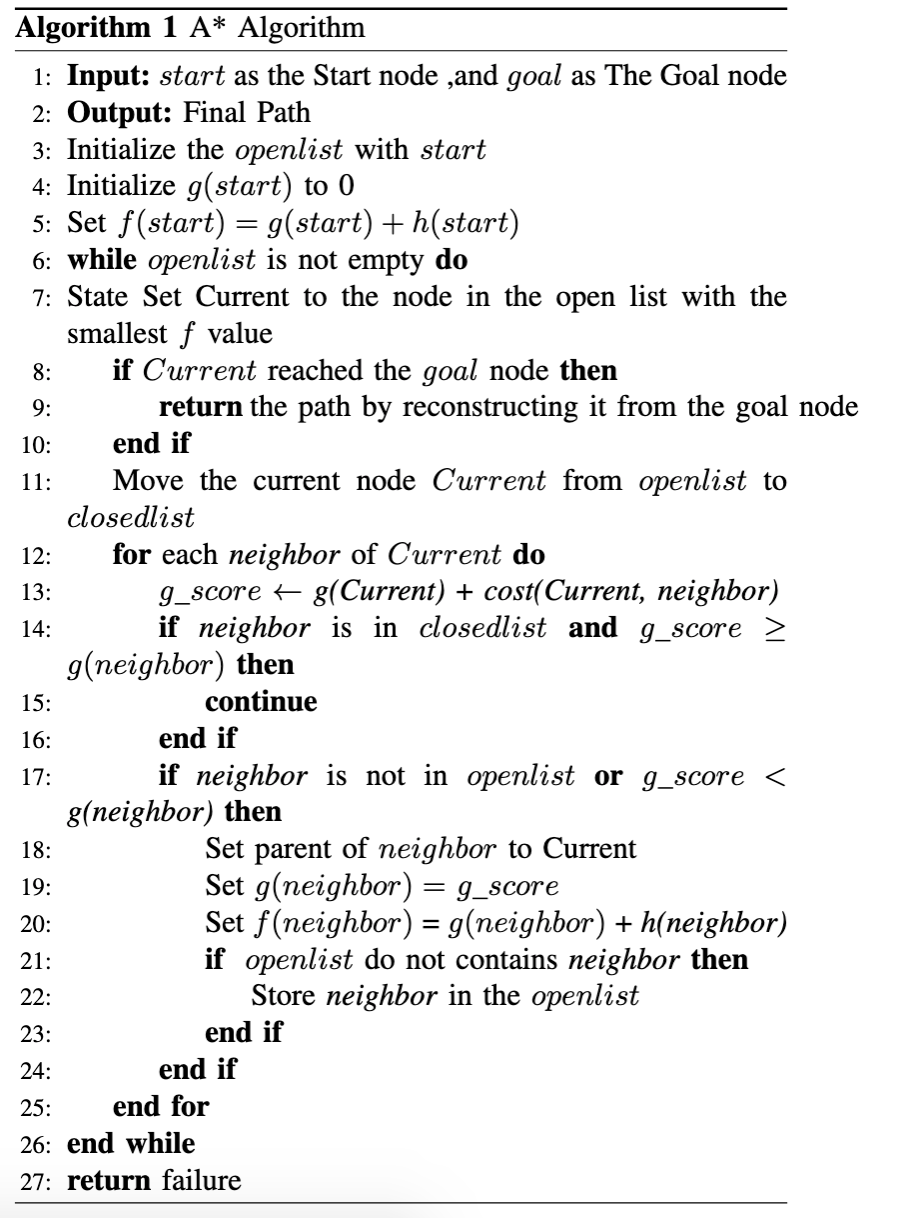}
    \caption{A* algorithm pseudo-code}
    \label{A*}
\end{figure}

The A* algorithm's pseudo code is outlined in Fig.\ref{A*}\\

Despite the simplicity of the A* algorithm, it has disadvantages. Firstly, the algorithm struggles in high-dimensional spaces and can be computationally expensive in execution time and storage due to the open and closed lists. In UAV path planning, the algorithm cannot guarantee the optimal path due to the limited number of each node's neighbors and the heuristic function \cite{b40}.

\subsection{RRT* Algorithm}
RRT* is a variation and an enhanced version of the Rapidly-exploring Random Tree (RRT) algorithm, designed to efficiently find optimal paths in high-dimensional spaces. The RRT algorithm is sampling-based, utilizing the search space to construct a random tree structure that begins from an initial position. It iteratively expands this tree by connecting each sampled point to the nearest node in the tree.

\begin{figure}
    \centering
    \includegraphics[width=1\linewidth]{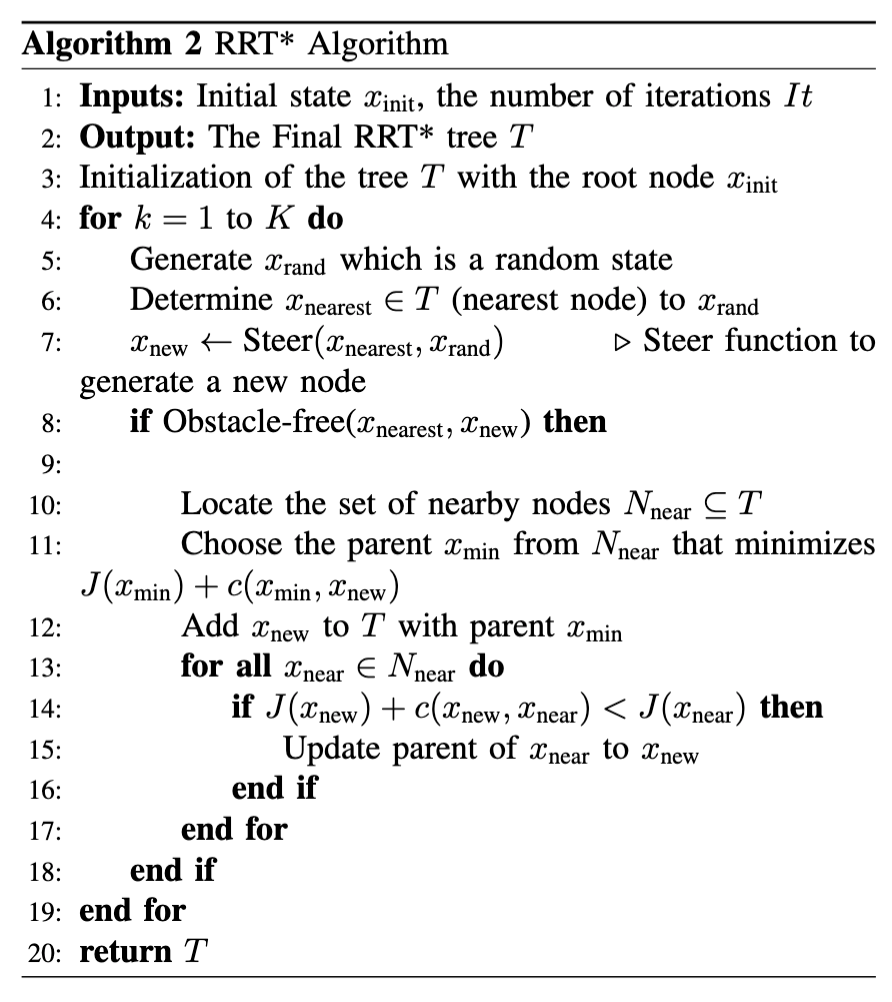}
    \caption{RRT* algorithm pseudo-code}
    \label{alg:rrt_star}
\end{figure}

Unlike RRT, which focuses on rapidly exploring the space and identifying any feasible path between the start and the target points, RRT* aims to discover the optimal path. Its basic idea is to use the existing tree and dynamically rewire it to improve path quality over time.
The algorithm starts with the starting point position. In each step, it randomly selects a new position in the search space to explore. It then finds the closest one in the path tree to this new position. The algorithm checks if this subpath collides with obstacles. If not, the newly created point is added to the tree, and the connection is returned to the closest existing node. Unlike basic RRT, RRT* also checks nearby nodes to see if they can find better routes through the new node. This process continually refines the tree structure to find the best possible path through the space \cite{b41}. 
The RRT* algorithm pseudo code is in Fig.\ref{alg:rrt_star}

RRT* is a popular algorithm for UAV path planning due to its effectiveness in navigating complex and high-dimensional environments. However, it has several limitations. The solution it finds is often only near-optimal due to the randomized sampling. Another limitation is that RRT* can find difficulty in finding feasible paths in environments with narrow channels. That means it depends on luck instead of calculations to find a path in these small areas \cite{b42}.

\subsection{Particle Swarm Optimization (PSO)}
PSO is a bio-inspired algorithm created in 1995 by J. Kennedy and R. Eberhart \cite{b9}. The algorithm utilizes a bio-inspired approach based on bird and fish foraging. PSO is widely used in path planning applications and can find the optimal route in different environments \cite{b43}. In PSO, the particles, which are a population of candidate solutions, navigate the search space to find the best or near-best solution. In each iteration, the particles update their positions based on two key factors: their own personal best positions ($pbest$) and the global best position ($pbest$) found within the entire population. This process continues iteratively until a specified number of iterations is reached. The quality of the solution is determined by a fitness function that evaluates each particle's position. This function, also known as the objective function, gives each particle a value, which represents its performance relative to the optimization goal. The fitness function is essential for guiding the swarm towards better solutions in the space. The equations used to adjust each particle's position and velocity are as follows:
 
 \begin{equation}
v_{i,d}^{t+1} = \omega v_{i,d}^t + c_1 r_1 (pbest_{i,d}^t - x_{i,d}^t) + c_2 r_2 (gbest_d^t - x_{i,d}^t) \label{eq:1}
\end{equation}

\begin{equation}
x_{i,d}^{t+1} = x_{i,d}^t + v_{i,d}^{t+1} \label{eq:2}
\end{equation}

\begin{figure}[h]
    \centering
    \includegraphics[width=1\linewidth]{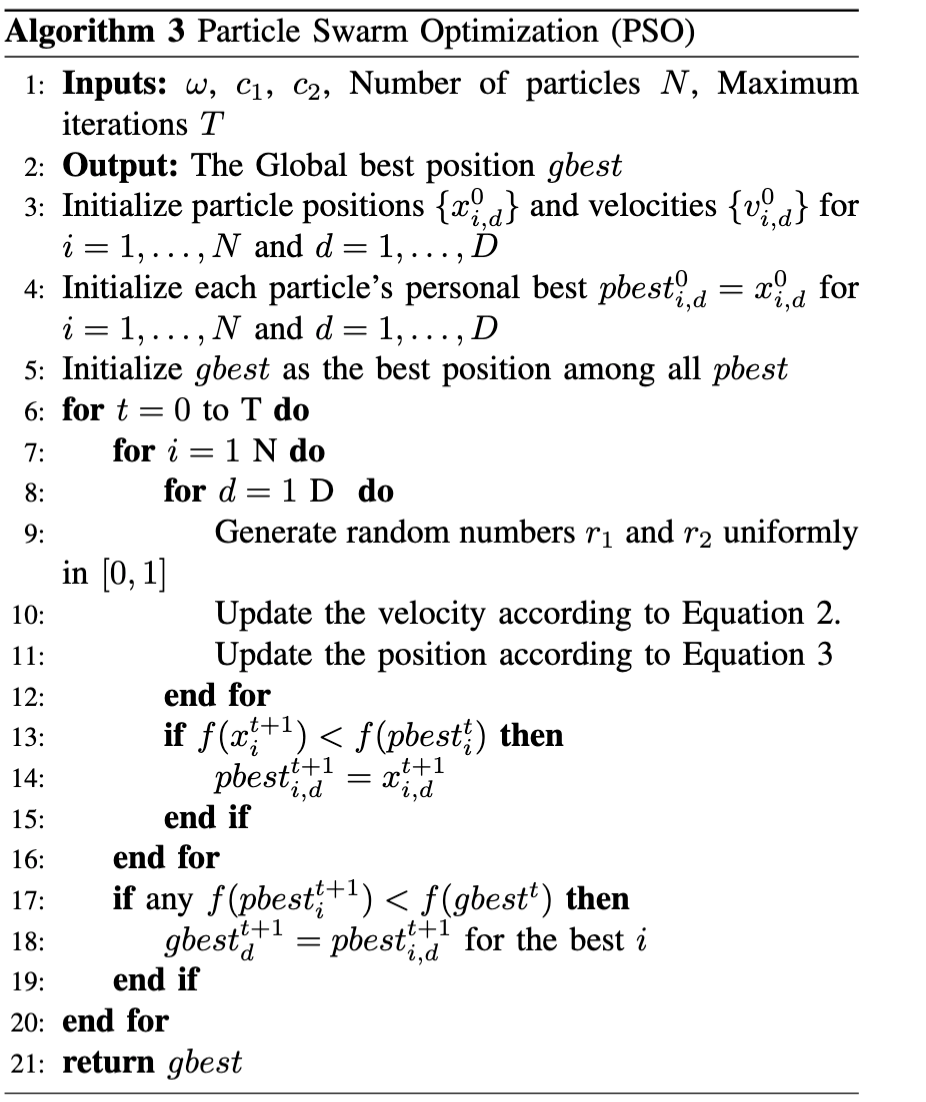}
    \caption{PSO algorithm pseudo-code}
    \label{alg:pso}
\end{figure}

Where \(v_{i,d}^{t}\) and \(v_{i,d}^{t+1}\) denote the \(i\)-th particle velocity at iteration \(t\) and \(t+1\) within \(d\) dimension respectively,\(\omega\) represents the inertia weight. \(pbest_{i,d}^t\) in the previous equation is the personel best position and \(gbest_{i,d}^t\) is the global best position of the \(i\)-th particle at iteration \(t\) and dimension \(d \). \(c1\) and \(c1\) are both learning factors. To be specific, \(c1\) is the self-learning factor that determines how much a particle relies on its individual experience when determining its future movements. Higher values of $c1$ mean that particles are increasingly influenced by their (\textit{pbest}). \(c2\) is the social learning factor, also referred to as the social acceleration factor, which represents a particle’s tendency to move toward the best solutions discovered by any member of the swarm. Its primary function is to draw the particle toward the (\textit{gbest}). \(r1\) and \(r2\) are a random numbers varying between \(0\) and \(1\), \(x_{i,d}^{t}\) represents the $i-th$ particle's current position, $d$ is for  dimension and $t$ for iteration number.
The RRT* algorithm pseudo code is in Fig.\ref{alg:pso}

Bio-inspired algorithms such as PSO are widely used in UAV path planning applications to find the shortest path from the start to the target point. However, these techniques may suffer from several limitations. Firstly, when planning a path in high-dimensional space, particles require significant computational resources and a large number of iterations. Secondly, due to the random movement of PSO particles, there is a risk of failing to find optimal solutions. Thirdly, The effectiveness of PSO is significantly influenced by its highly sensitive parameters. These selected parameters include population size, inertia weight, and both cognitive and social coefficients. Finding the right combination of parameters is challenging and requires multiple experiments \cite{b44}.

\section{SIMULATIONS AND ANALYSIS}

The following experiment and algorithm were conducted on a MacBook Pro 2020 with an i5 processor and 8GB RAM, running MATLAB 2023a.

\subsection{Environment Modelling}
In a real-world 3D environment, UAVs encounter various challenges. For instance, they must sometimes avoid a single obstacle and, at other times, multiple obstacles simultaneously. Additionally, UAVs may need to operate at varying altitudes, either high or low. To evaluate the effectiveness of the three algorithms and make a comparative analysis, this paper employs 3 distinct experiments with two different scenarios each.

\subsubsection{Experiment 1: Obstacle Density Variation}
The first scenario sets obstacles at a high density of 60\%, which creates a densely urban environment intended to challenge pathfinding algorithms with numerous obstacles. Conversely, the second scenario reduces obstacle density to 10\%.This results in a less challenging environment with more open spaces and fewer obstacles. These scenarios aim to evaluate algorithm performance across different urban conditions and highlight their adaptability and efficiency in navigating varying levels of spatial complexity.

\subsubsection{Experiment 2: Travel Distance Variation}
This experiment investigates how three algorithms perform under varying travel distances across maps of different sizes. In Scenario 3, the algorithms are tested over a long travel distance between the start and endpoint on a large city map of 2 km by 2 km. This scenario evaluates their efficiency, reliability, and ability to maintain optimal pathfinding strategies over extended distances. Conversely, Scenario 4 tests the algorithms over a short travel distance on a smaller map of 1 km by 1 km, which challenges the three algorithms with shorter navigation tasks. 

\subsubsection{Experiment 3: Altitude Variation}
This experiment evaluates three algorithms' performance in navigating urban environments with different altitude conditions. Scenario 5 tests the algorithms with significant altitude differences between the start and endpoint. This challenges their pathfinding abilities over diverse urban terrains. Meanwhile, Scenario 6 examines their navigation capabilities when the start and endpoint are at the same altitude level within the city.

In addition, several constraints must be considered when testing three algorithms for path-planning missions. Firstly, to ensure safe collision avoidance, this paper sets a security distance of 1 m between the generated path and obstacles. Moreover, in Experiment 3, the algorithms are tested at different altitudes, with a maximum altitude difference of 30 m between the start and target points. Additionally, this paper considers battery life constraints by using a limited flying range of 200 m for all scenarios except Scenario 4 (2 km x 2 km), where the maximum range is 400 m. Furthermore, any sharp turns less than 30 degrees are considered unsuitable for the planned path and will not be considered. This paper uses the Euclidean distance for the A* algorithm's heuristic function. For the RRT* algorithm, the random choosing rate is set to 0.5. For the PSO algorithm, the paper selects 200 for the iteration number, 150 for the population size, and 1.9 for both c1 and c2 learning factors.\\

\subsection{Evaluation Metrics}
To effectively test the three algorithms' efficacy and performance, three objective functions have been adopted, including path distance, number of turning angles, and algorithm running time.\\

\subsubsection{Path Length $P$}
While path length directly reduces travel time, choosing the shortest path is essential for the path planning algorithm’s objective function. The following function is used to calculate the path length $P$:
\begin{equation}
L = \sum_{i=1}^{n-1} \sqrt{(x_{i+1} - x_i)^2 + (y_{i+1} - y_i)^2 + (z_{i+1} - z_i)^2}
\end{equation}
where $n$ denotes the number of nodes, $x_i$, $y_i$, and $z_i$ are the coordinates of each node $i$ in the path.\\

\subsubsection{Number Of Turning Angles}
The number of turning angles significantly impacts UAV path efficiency, stability, and maneuverability. Each turn affects path length, flight stability, and the UAV's ability to navigate obstacles and execute mission objectives effectively. The following formula is used:

\begin{align}
T_r(\theta)& = \sum_{i=1}^{n-1} \theta_i. \\
\vec{a} &= (x_{n} - x_{n-1}, y_{n} - y_{n-1}, z_{n} - z_{n-1})\\
\vec{b} &= (x_{n+1} - x_{n}, y_{n+1} - y_{n}, z_{n+1} - z_{n})\\
\cos \theta_n &= \frac{\vec{a} \cdot \vec{b}}{\|\vec{a}\| \|\vec{b}\|} 
\end{align}

Here, \( n \) represents the \( n \)-th path node, and \( \theta_n \) is the angle at the \( n \)-th node, calculated based on the positions of the nodes \( n-1 \), \( n \), and \( n+1 \). \(\vec{a}\) and \(\vec{b}\) are two vectors calculated based on the \( n-1 \), \( n \), and \( n+1 \) path nodes. \( |\vec{a}| \) and \( |\vec{b}| \) are the magnitudes of \(\vec{a}\) and \(\vec{b}\), respectively.

\subsubsection{Algorithm Time}
the execution time of the algorithm strongly influences the overall mission operation time. Therefore, an objective function is defined as follows:

\begin{equation}
    T_{alg} = toc - tic
\end{equation}

Where $toc$ and $tic$ are predefined instructions to calculate the time of the algorithm using Matlab Toolbox

\subsection{Experiment Analysis And Comparison}

\begin{figure*}
    \centering
    \subfloat[A* Algorithm]{
        \includegraphics[width=0.30\linewidth]{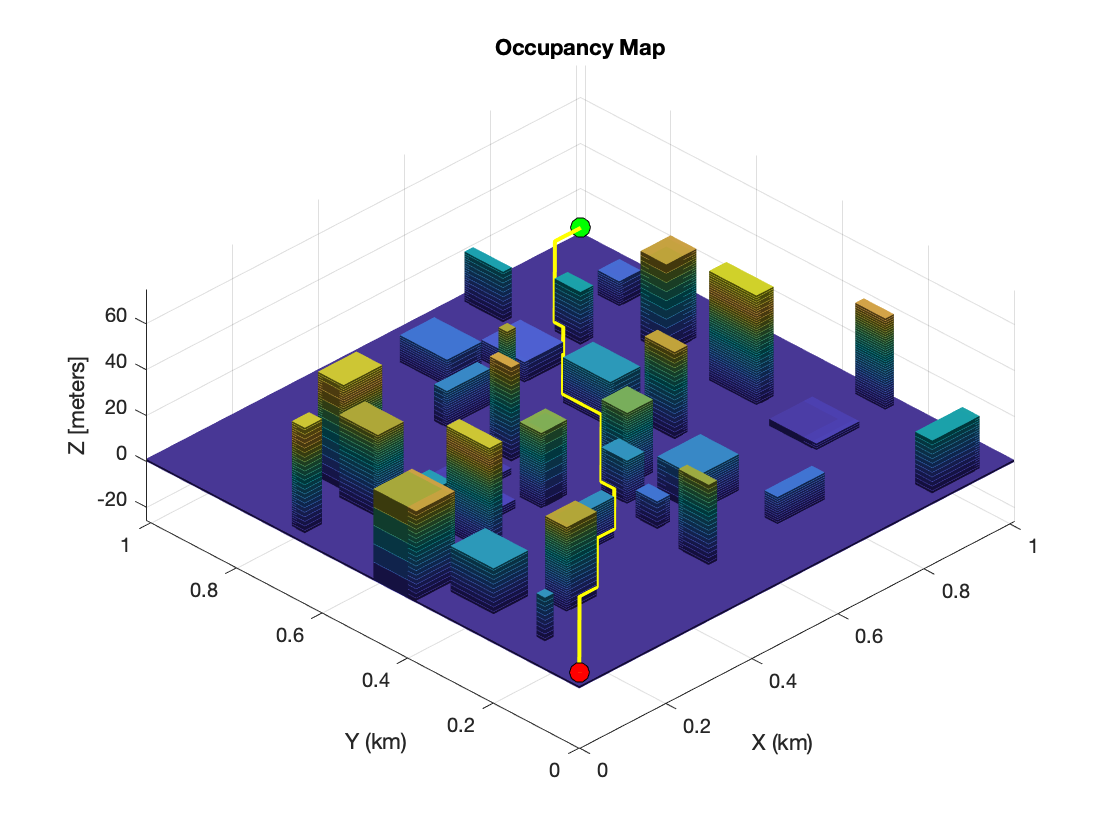}
        \label{fig:1A*}
    }
    \hfill
    \subfloat[RRT* Algorithm]{
        \includegraphics[width=0.30\linewidth]{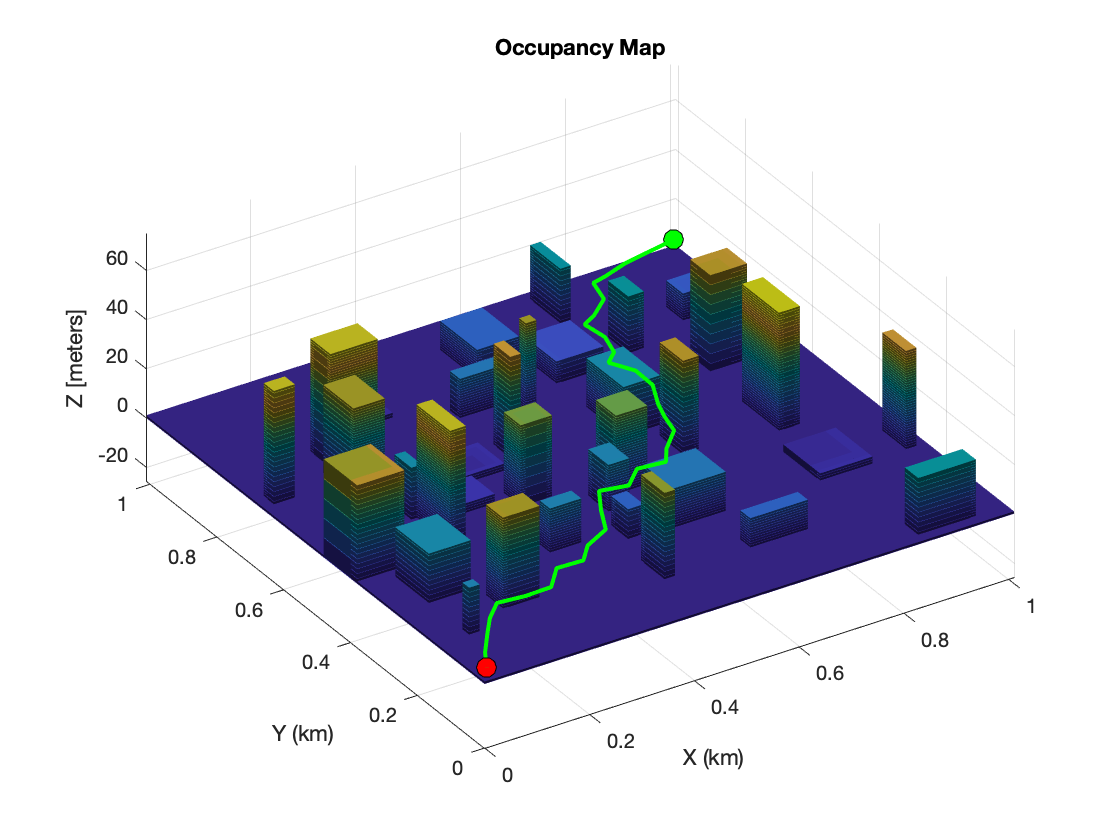}
        \label{fig:1RRT*}
    }
    \hfill
    \subfloat[PSO Algorithm]{
        \includegraphics[width=0.30\linewidth]{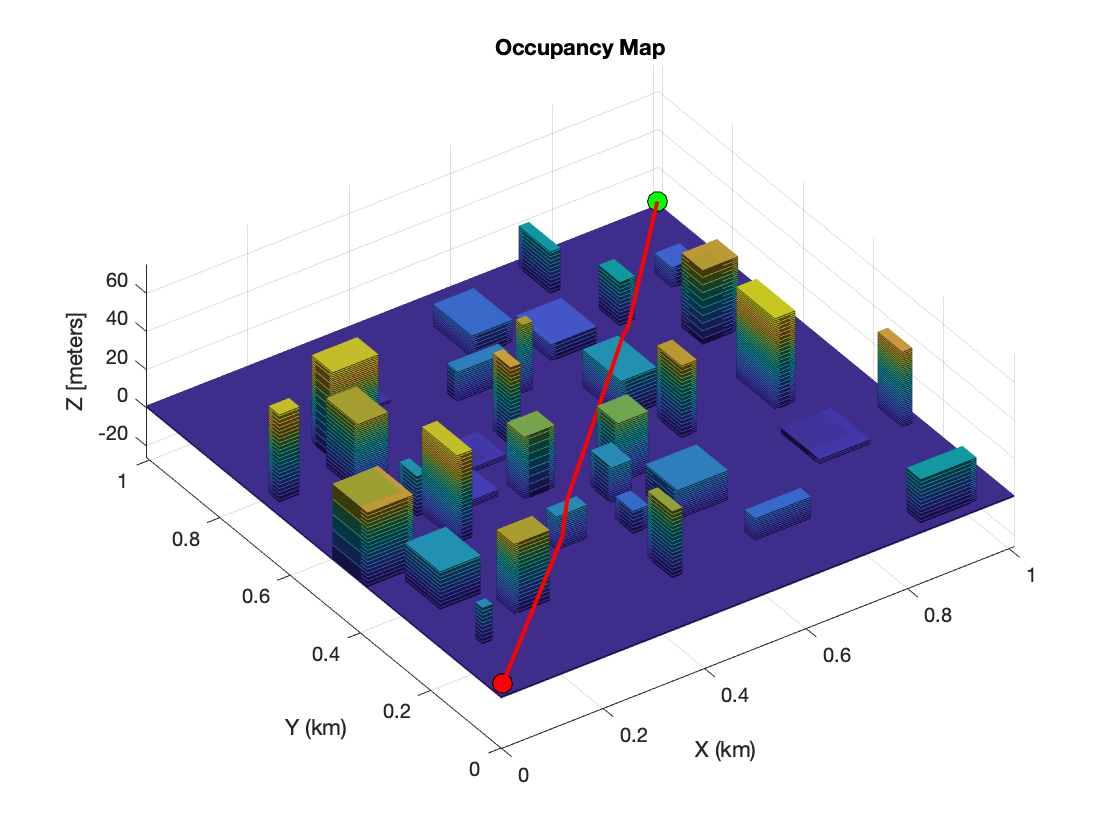}
        \label{fig:1PSO}
    }

    \caption{Generated Paths Using A*, RRT* and PSO Algorithms}
    \label{fig:A*RRT*PSO}
\end{figure*}

As shown in Fig.\ref{fig:1A*}, The A* algorithm always chooses the lowest cost path, which makes the UAV fly close to obstacles and at low altitudes. However, this path has many turns, making UAVs consume more energy. Conversely, the planned path using the RRT* algorithm, shown in Fig.\ref{fig:1RRT*}, has even more turns due to its random sampling strategy. Some of these turns are sharp, adversely affecting the UAV's dynamics. Finally, as seen in Fig.\ref{fig:1PSO}, the path generated by the PSO algorithm forms approximately a straight line connecting the start and the goal points. This planned trajectory allows the UAV to make small angle adjustments, resulting in lower energy consumption.

\begin{figure*}[!h]
    \centering
    \subfloat[Less Obstacles]{
        \includegraphics[width=0.30\linewidth]{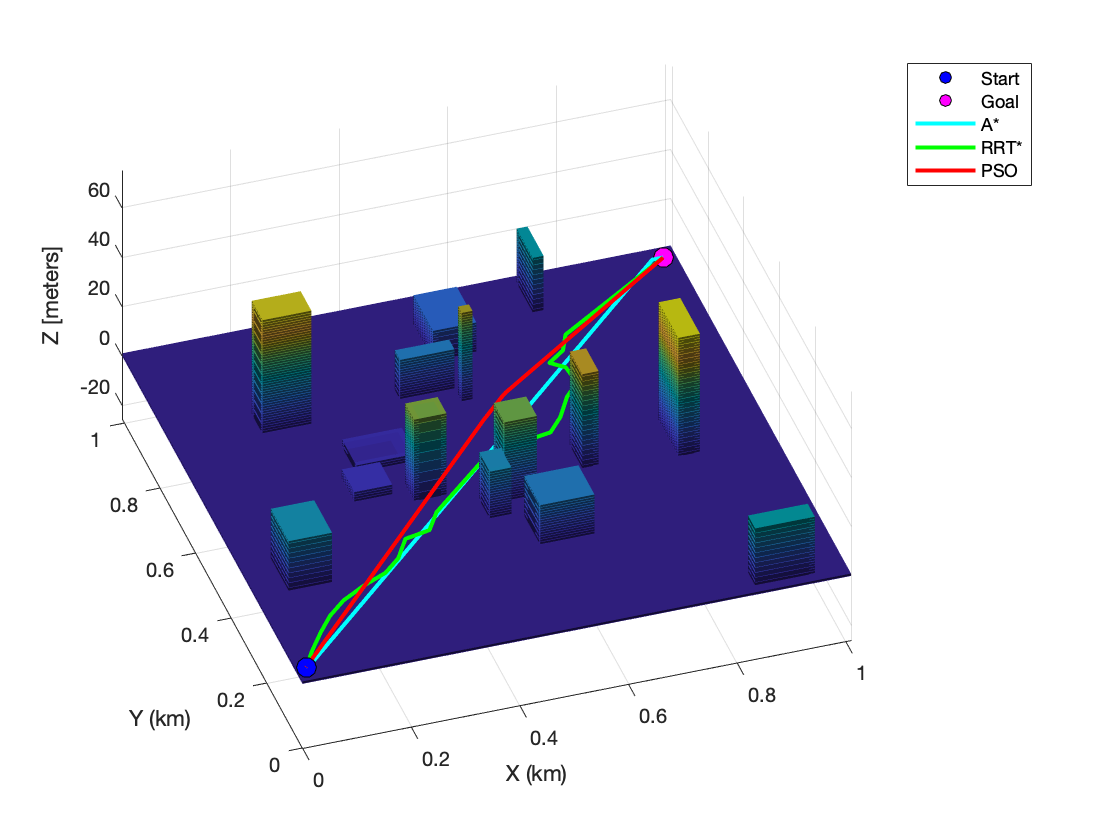}
        \label{fig:less}
    }
    \hfill
    \subfloat[Dense Obstalces]{
        \includegraphics[width=0.30\linewidth]{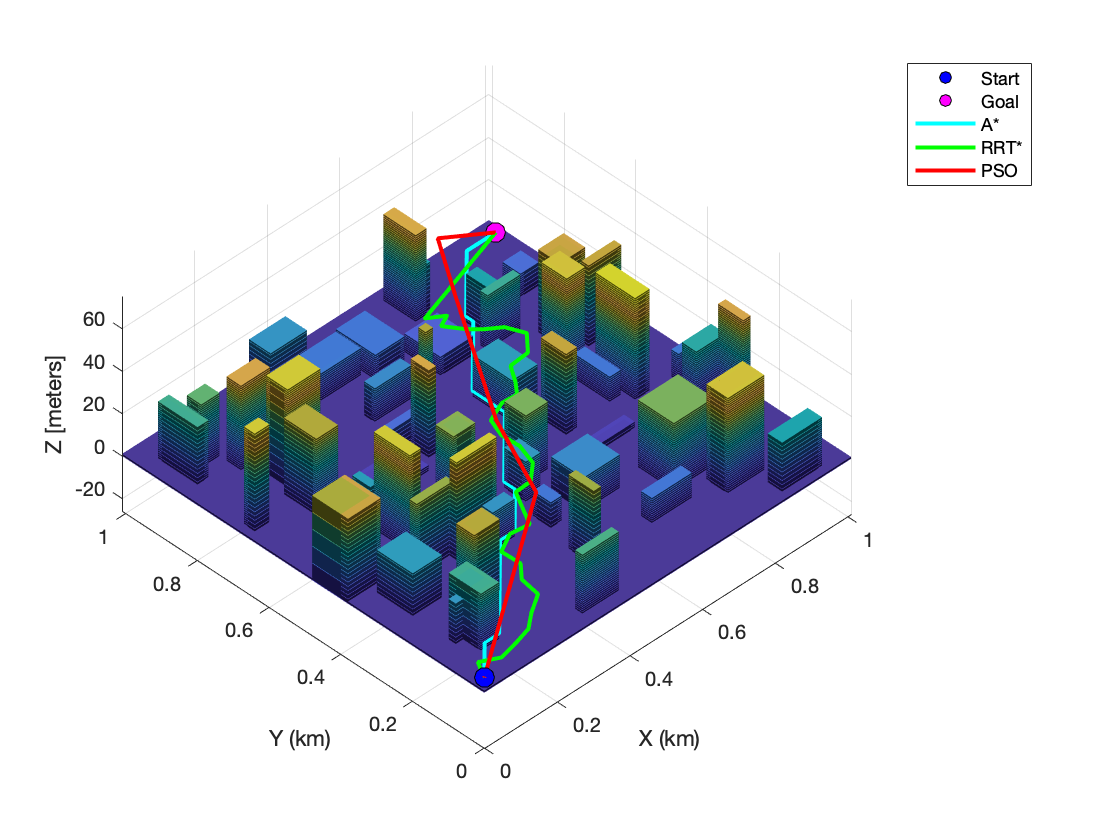}
        \label{fig:dense}
    }
    \hfill
    \subfloat[Large Altitude Difference]{
        \includegraphics[width=0.30\linewidth]{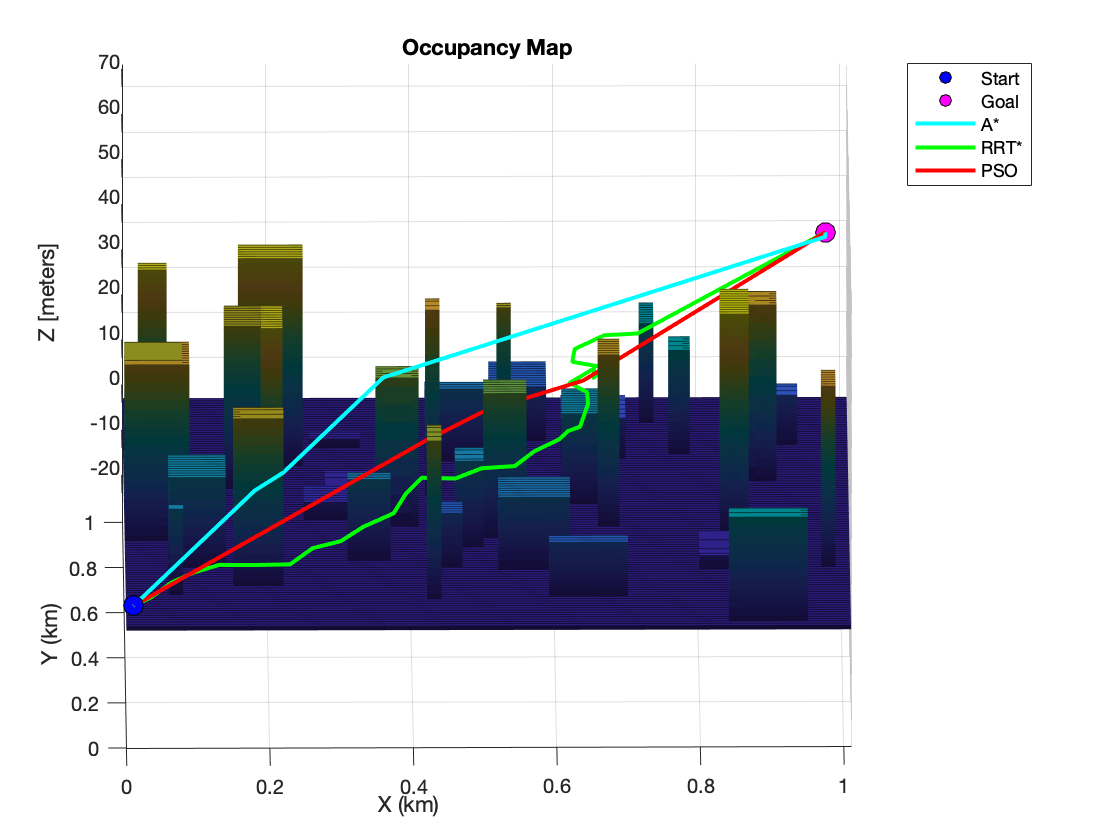}
        \label{fig:large_altitude}
    }

    \hfill
    \subfloat[Short Altitude Difference]{
        \includegraphics[width=0.30\linewidth]{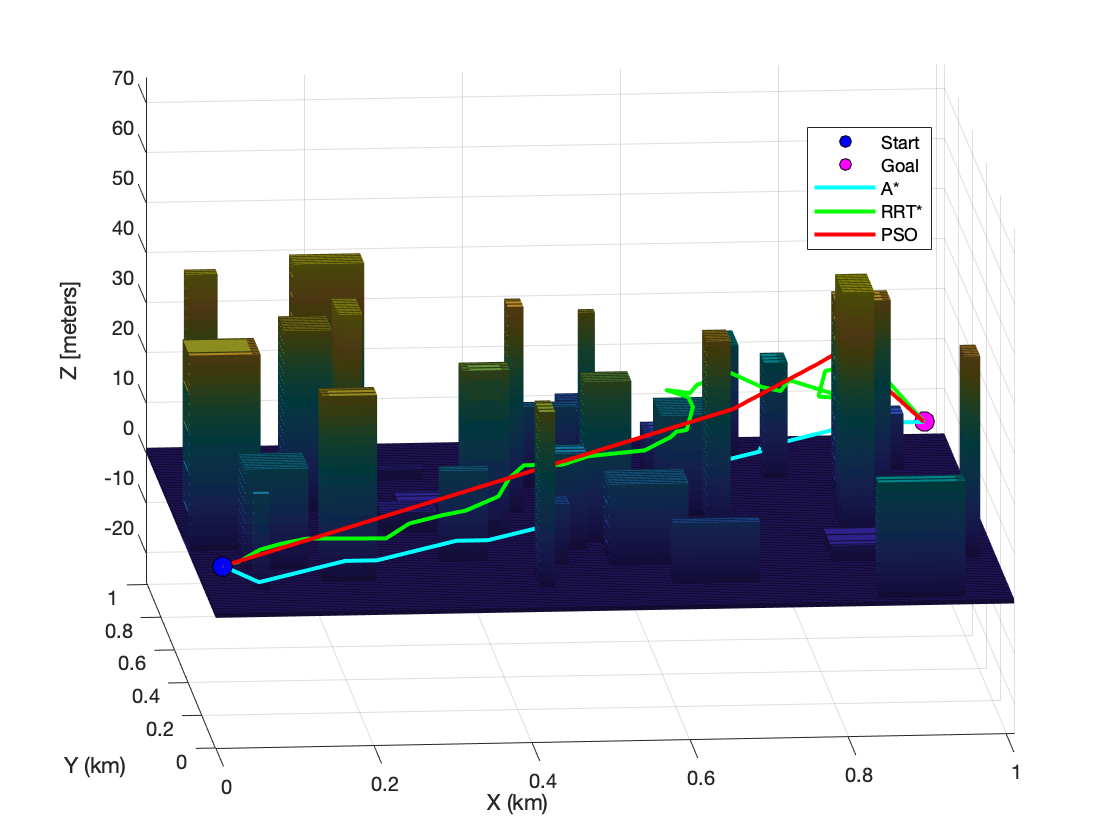}
        \label{fig:small_altitude}
    }
    \hfill
    \subfloat[Large Map]{
        \includegraphics[width=0.30\linewidth]{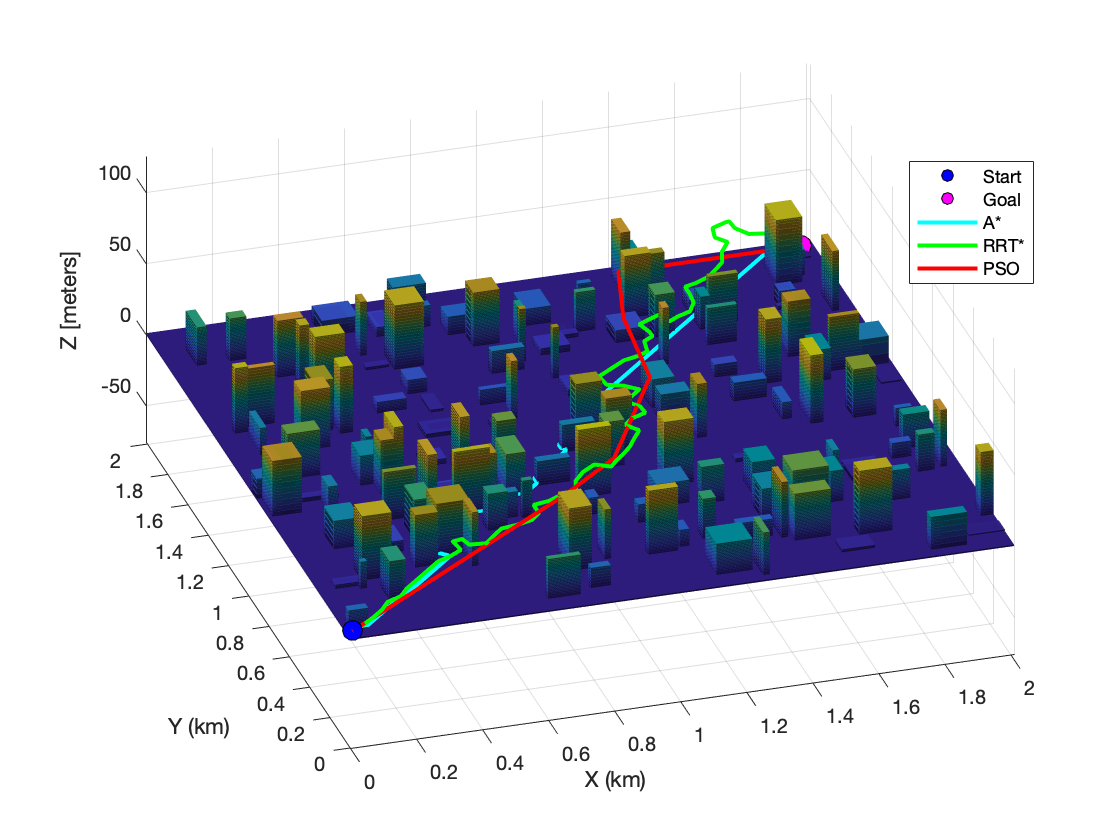}
        \label{fig:large_map}
    }
    \hfill
    \subfloat[Small Map]{
        \includegraphics[width=0.30\linewidth]{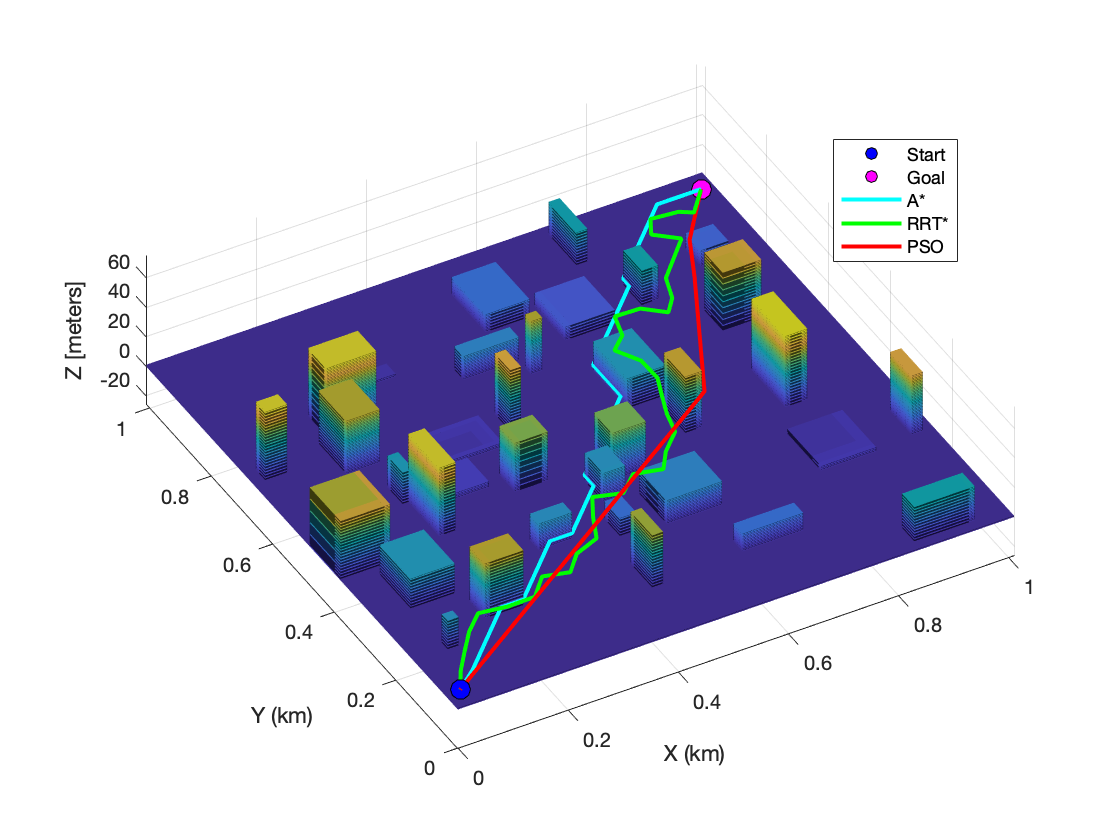}
        \label{fig:small_map}
    }

    \caption{Comparisons of Generated Paths Under Different Scenarios}
    \label{fig:pathcomparison}
\end{figure*}

\begin{figure*}[!h]
    \centering
    \subfloat[Path Length]{
        \includegraphics[width=0.3\linewidth]{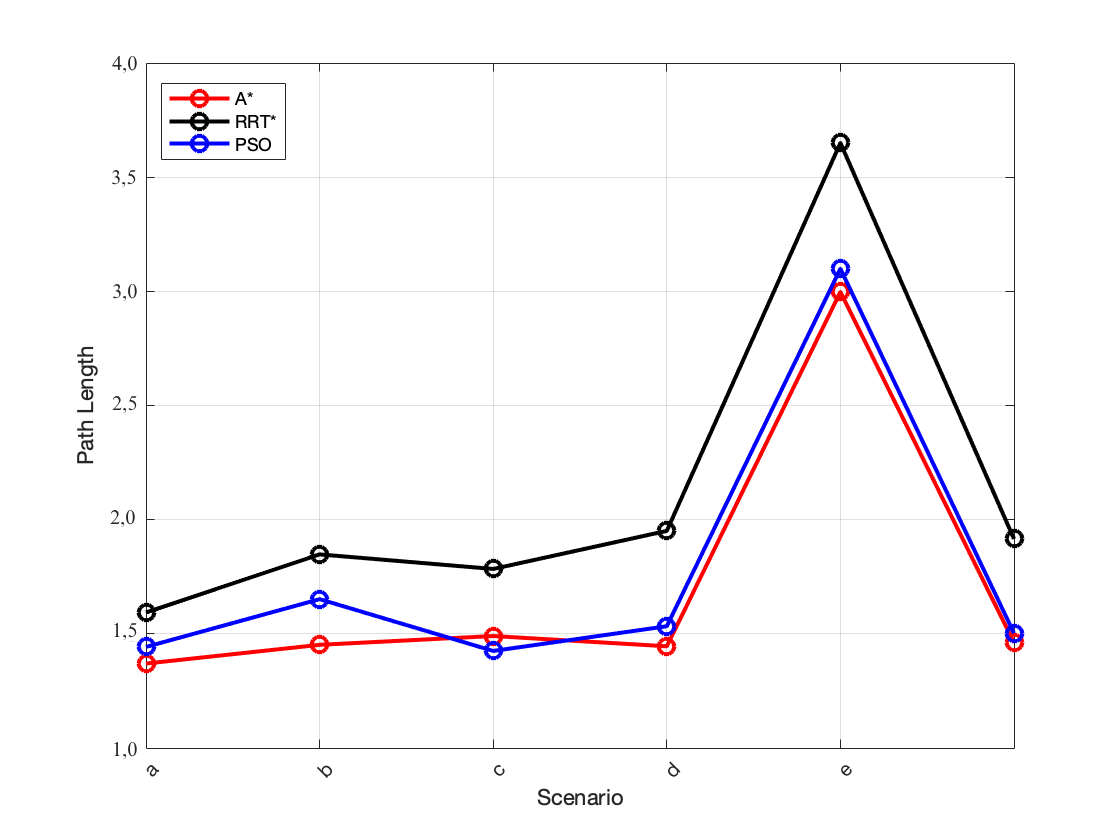}
        \label{fig:pathlength}
    }
    \hfill
    \subfloat[Turning Angles]{
        \includegraphics[width=0.3\linewidth]{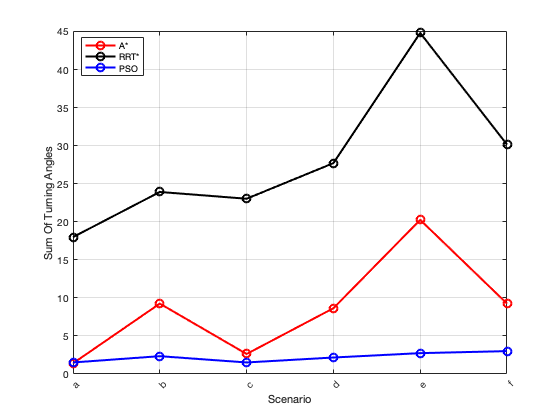}
        \label{fig:turningangles}
    }
    \hfill
    \subfloat[Algorithm Time]{
        \includegraphics[width=0.3\linewidth]{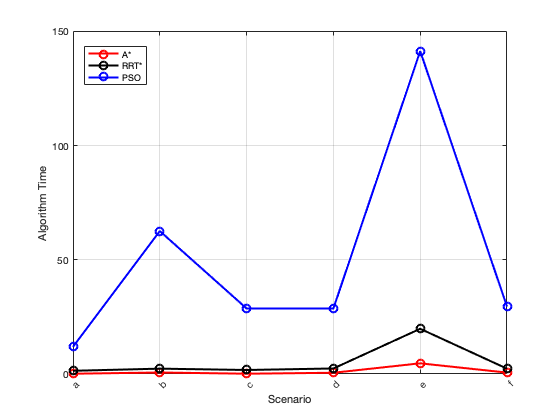}
        \label{fig:algotime}
    }

    \caption{Comparison of various metrics (a) Path Length, (b) Turning Radius, and (c) Algorithm Computation Time across different algorithms and map scenarios.}
    \label{fig:comparison}
\end{figure*}

As shown in Fig. \ref{fig:pathcomparison}, three paths generated by A*, RRT*, and PSO algorithms are compared under different conditions. Fig. \ref{fig:less} and Fig. \ref{fig:dense} represent Experiment 1, which tests path planning under varying obstacle densities. We observe that the paths planned by the three algorithms in Fig. \ref{fig:less} are shorter, have fewer turning points, and maintain a lower altitude range compared to the paths generated in Fig. \ref{fig:dense}. Experiment 2, represented by Fig.\ref{fig:large_altitude} and Fig.\ref{fig:small_altitude}, test the algorithms under different altitudes. It demonstrates that the A* algorithm and PSO produce nearly straight lines, while RRT* continues to generate random nodes at lower altitudes.  Unlike Experiment 1, The A* uses the absence of obstacles at a high altitude to minimize the turns. Fig.\ref{fig:large_map} shows the results of Experiment 3, which is described in the previous section. From the results in Fig.\ref{fig:large_map}, it is obvious that generated paths are longer compared to Fig.\ref{fig:small_map} because of the size of the map.

Fig.\ref{fig:comparison} shows the comparison of path length (km), the sum of turning angles (rad), and the algorithm time (s) under the same scenarios. From Fig.\ref{fig:pathlength}, we can see that the A* algorithm consistently produces the shortest paths, outperforming RRT* and PSO. On average, the path length for A* is about 20\% shorter than RRT* and around 5\% shorter than PSO. Additionally, both the map's size and the density of obstacles significantly impact the path length.

When examining the sum of turning angles in Fig.\ref{fig:turningangles}, particularly in handling tight turns. On average, PSO's turning radius is approximately 85\% smaller than that of RRT* and about 70\% smaller than that of A*. For example, in map scenario 'f,' PSO achieves a turning radius of 3 rad, which is 90\% smaller than RRT*'s turning radius of 30.13 rad and approximately 67\% smaller than A*'s turning radius of 9.25 rad. This significant reduction in turning radius indicates that PSO is more adept at navigating through tighter spaces than RRT* and A*.

Regarding computation times, A* is the most efficient algorithm, often completing its calculations in a fraction of the time taken by RRT* and PSO. A* is typically 90\% faster than PSO and about 75\% faster than RRT*. For instance, in map scenario 'e', A* completes in 4.6759 seconds, which is approximately 76\% faster than RRT*'s 19.7987 seconds and a remarkable 97\% faster than PSO's 141.3934 seconds.

\section{Conclusion}
From the findings discussed above, it is evident that the A* is the preferred algorithm for UAV path planning due to its optimal path length and rapid computation time, making it highly effective for navigation. PSO's ability to handle tight turns and dense environments makes it a valuable algorithm in specific scenarios, but its high computation time is a drawback. RRT* offers a balance but falls short in comparison to A*'s overall efficiency due to its randomness.

\section{Declaration of intersets}
The authors report no conflicts of interest relevant to this study.

\section{Data availability}
Data can be obtained upon request.

\section*{Acknowledgment}

This work was supported by the research project "Modeling and Control of Aerial Manipulators" N° C00L07UN310220230004.

\end{document}